\documentclass[11pt,a4paper]{article}
\usepackage[hyperref]{acl2019}
\usepackage{times}
\usepackage{latexsym}
\usepackage{graphicx}
\usepackage{url}
\usepackage{amsmath}
\usepackage{amssymb}
\usepackage{empheq}
\usepackage{framed}
\usepackage{array}

\newif\ifcomments
\commentstrue 
\ifcomments
    \newcommand{\BB}[1]{{\color{red}(BB: #1)}}
    \newcommand{\jb}[1]{{\color{blue}(JB: #1)}}
    \newcommand{\matt}[1]{{\color{olive}(MG: #1)}}
\else
    \providecommand{\BB}[1]{}
    \providecommand{\jb}[1]{}
    \providecommand{\matt}[1]{}
\fi

\newcommand{\sT}{\mathcal{T}}
\newcommand{\sC}{\mathcal{C}}

\newcommand{\sF}{\mathcal{F}}
\newcommand{\sV}{\mathcal{V}}
\newcommand{\sE}{\mathcal{E}}

\newcommand{\vlegal}{\sV_{\text{legal}}}
\newcommand{\glegal}{G_{\text{legal}}}
\newcommand{\softmax}{\text{softmax}}
\newcommand{\spider}{\textsc{Spider}}

\newcommand{\correct}[1]{{\color{blue}#1}}
\newcommand{\notcorrect}[1]{#1}

\aclfinalcopy 

\title{Representing Schema Structure with Graph Neural Networks for Text-to-SQL Parsing}

\author{Ben Bogin$^{1}$ ~~~~~ Matt Gardner$^{2}$ ~~~~~
Jonathan Berant$^{1,2}$ \\
\mbox{}\\
$^1$School of Computer Science, Tel-Aviv University \\
$^2$Allen Institute for Artificial Intelligence \\
\small{\texttt{ben.bogin@cs.tau.ac.il,mattg@allenai.org,joberant@cs.tau.ac.il}}}

\date{}

\begin{document}

\maketitle

\begin{abstract}
Research on parsing language to SQL has largely ignored the structure of the database (DB) schema, either because the DB was very simple,
or because it was observed at both training and test time. In \spider{}, a recently-released text-to-SQL dataset, new and complex DBs are given at test time, and so the structure of the DB schema can inform the predicted SQL query. In this paper, we present an encoder-decoder semantic parser, where the structure of the DB schema is encoded with a graph neural network, and this representation is later used at both encoding and decoding time. 
Evaluation shows that encoding the schema structure improves our parser accuracy  from 33.8\% to 39.4\%, dramatically above the current state of the art, which is at 19.7\%.
\end{abstract}

\section{Introduction}

Semantic parsing \cite{zelle96geoquery,zettlemoyer05ccg} has recently taken increased interest in parsing questions into SQL queries, due to the popularity of SQL as a query language for relational databases (DBs).

Work on parsing to SQL  \cite{zhong2017seq2sql,iyer2017neural,finegan2018improving,Yu2018SyntaxSQLNet} has either involved simple DBs that contain just one table, or had a single DB that is observed at both training and test time. Consequently, modeling the \emph{schema structure} received little attention. Recently, \newcite{yu2018spider} presented \spider{}, a text-to-SQL dataset, where
at test time questions are executed against unseen and complex DBs. In this zero-shot setup, an informative representation of the schema structure is important. Consider the questions in Figure~\ref{fig:importance}: while their language structure is similar,
in the first query a `join' operation is necessary because the information is distributed across three tables, while in the other query no `join' is needed.

In this work, we propose a semantic parser that strongly uses the schema structure. We represent the structure of the schema as a graph, and use graph neural networks (GNNs) to provide a 
global representation for each node
\cite{li2016gated,decao2018question,sorokin2018modeling}. 
We incorporate our schema representation into the encoder-decoder parser of \newcite{krishnamurthy2017neural}, which was designed to parse questions into queries against unseen semi-structured tables. At encoding time we enrich each question word with a representation of the subgraph it is related to, and at decoding time we emit symbols from the schema that are related through the graph to previously decoded symbols.

We evaluate our parser on \spider{}, and show
that encoding the schema structure improves accuracy from 33.8\% to 39.4\% (and from 14.6\% to 26.8\% on questions that involve multiple tables), well beyond 19.7\%, the current state-of-the-art. We make our code publicly available at \url{https://github.com/benbogin/spider-schema-gnn}.

\begin{figure}
    \scriptsize
    \begin{framed}
        $x:$ \textit{\color{blue}{Find the age of students who \textbf{do not have} a cat pet.}}
        \\
        $y:$ \texttt{SELECT age FROM \underline{student} WHERE \newline student NOT IN \color{red}{(SELECT ... FROM \underline{student} \textbf{JOIN} \underline{has\_pet} ... \textbf{JOIN} \underline{pets} ... WHERE ...)}} \\
        $x:$ \textit{\color{blue}{What are the names of teams that \textbf{do not have} match season record?}} \\~
        $y:$ \texttt{SELECT name FROM \underline{team} WHERE \newline team\_id NOT IN \color{red}{(SELECT team FROM \underline{match\_season})}}
        \end{framed}
    \caption{Examples from \spider{} showing how similar questions can have different SQL queries, conditioned on the schema. Table names are underlined.}
    \label{fig:importance}
\end{figure}

\section{Problem Setup}
We are given a training set $\{(x^{(k)}, y^{(k)}, S^{(k)})\}_{k=1}^N$, where $x^{(k)}$ is a natural language question, $y^{(k)}$ is its translation to a SQL query, and $S^{(k)}$ is the schema of the DB where $y^{(k)}$ is executed. Our goal is to learn a function that maps an unseen question-schema pair $(x,S)$ to its correct SQL query. Importantly, the schema $S$ was not seen at training time, that is, $S \neq S^{(k)}$ for all $k$. 

A DB schema $S$ includes: (a) The set of DB tables $\sT$ (e.g., \texttt{singer}), (b) a set of columns $\sC_t$ for each $t \in \sT$ (e.g., \texttt{singer\_name}), and (c) a set of foreign key-primary key column pairs $\sF$, where each $(c_f, c_p) \in \sF$ is a relation from a foreign-key $c_f$ in one table to a primary-key $c_p$ in another. We term all schema tables and columns as \emph{schema items} and denote them by $\sV = \sT \cup \{\sC_t\}_{t \in \sT}$. 

\section{A Neural Semantic Parser for SQL}
\label{sec:base_model}

\begin{figure}
    \centering
    \includegraphics[scale=0.4]{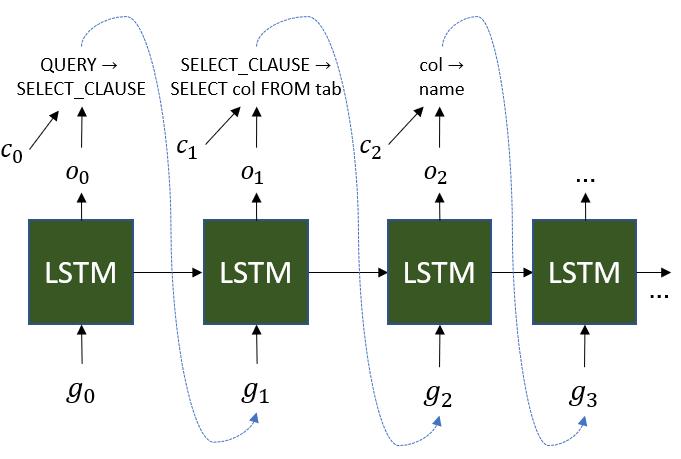}
    \caption{The decoder we base our work on \cite{krishnamurthy2017neural}. The input to the LSTM ($g_j$) at step $j$ is a learned embedding of the last decoded grammar rule, except when the last rule is schema-specific ($g_3$), where the input is a learned embedding of the schema item type. A grammar rule is selected based on the LSTM output ($o_j$) and the attended hidden state of the input LSTM ($c_j$). }
    \label{fig:base_decoder}
\end{figure}

We base our model on the parser of \newcite{krishnamurthy2017neural}, along with a grammar for SQL provided by AllenNLP \cite{gardner2018allennlp,lin2019grammar}, which covers 98.3\% of the examples in \spider{}. This parser uses a linking mechanism for handling unobserved DB constants at test time. We review this model in the context of text-to-SQL parsing, focusing on components we expand upon in \S\ref{sec:model}.

\paragraph{Linking schema items} 
To handle unseen schema items, \newcite{krishnamurthy2017neural} learn a similarity score $s_\text{link}(v, x_i)$ between a word $x_i$ and a schema item $v$ that has type $\tau$.\footnote{Types are tables, string columns, number columns, etc.} The score is based on learned word embeddings and a few manually-crafted features. 
 
The linking score is used to compute
$$p_\text{link}(v \mid x_i) = \frac{\exp (s_\text{link}(v,x_i))}{\sum_{v'\in \sV_\tau \cup \{\varnothing\}} \exp (s_\text{link}(v',x_i))},$$
where $\sV_\tau$ are all schema items of type $\tau$ and $s_\text{link}(\varnothing, \cdot)=0$ for words that do not link to any schema item. The functions $p_\text{link}(\cdot)$ and $s_\text{link}(\cdot)$ will be used to decode unseen schema items.

\paragraph{Encoder} A Bidirectional LSTM \cite{hochreiter1997lstm} provides a contextualized representation $h_i$ for each question word $x_i$.
Importantly, the encoder input at time step $i$ is $[w_{x_i} ; l_i]$: the concatenation of the word embedding for $x_i$ and  $l_i = \sum_\tau \sum_{v \in \sV_\tau} p_\text{link}(v \mid x_i) \cdot r_v $, where $r_v$ is a learned embedding for the schema item $v$, based on the \emph{type} of $v$ and its schema neighbors.
Thus, $p_\text{link}(v \mid x_i)$ augments every word $x_i$ with information on the schema items it should link to.

\paragraph{Decoder}
We use a grammar-based \cite{xiao2016sequence,cheng2017learning,yin2017syntactic,rabinovich2017abstract} LSTM decoder with attention on the input question (Figure \ref{fig:base_decoder}). 
At each decoding step, a non-terminal of type $\tau$ is expanded using one of the grammar rules. Rules are either schema-independent and generate
non-terminals or SQL keywords, or schema-specific and generate schema items. 

At each decoding step $j$, the decoding LSTM takes a vector $g_{j}$ as input, which is an embedding of the grammar rule decoded in the previous step, and outputs a vector $o_j$. If this rule is schema-independent, $g_{j}$ is a learned global embedding. If it is schema-specific, i.e., a schema item $v$ was generated, $g_{j}$ is a learned embedding $\tau(v)$ of its type. An attention distribution $a_j$ over the input words is computed in a standard manner \cite{bahdanau2015neural}, where the attention score for every word is $h_i^\top o_j$. It is then used to compute the weighted average of the input $c_j = \sum_i a_j h_j$. Now a distribution over grammar rules is computed by:
\begin{align*}
    s^\text{glob}_j &= \text{FF}(\big[o_j; c_j \big])  \quad \in \mathbb{R}^{\glegal}, \\
    s^\text{loc}_j &= S_\text{link} a_j \quad \in \mathbb{R}^{\vlegal}, \\
    p_j &= \softmax([s^\text{glob}_j; s^\text{loc}_j]),
\end{align*}

\begin{figure*}
    \centering
    \includegraphics[scale=0.3]{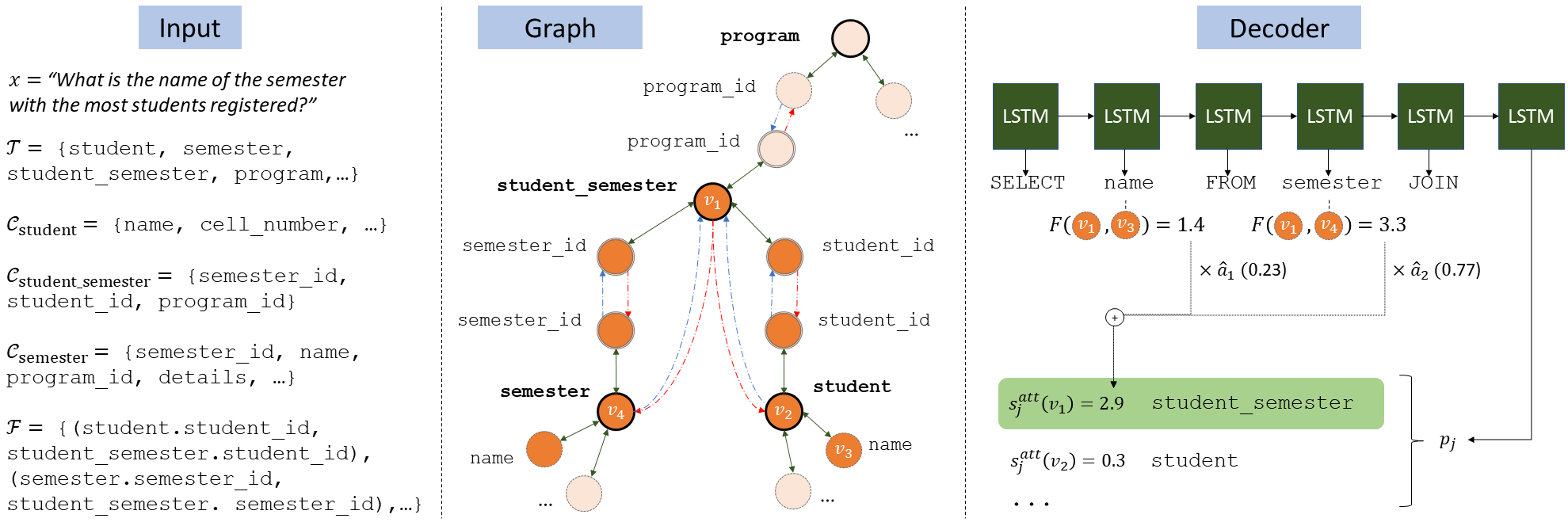}
    \caption{\underline{Left}: DB schema and question. \underline{Middle}: A graph representation of the schema. Bold nodes are tables, other nodes are columns. Dashed red (blue) edges are foreign (primary) keys edges, green edges are table-column edges.
    \underline{Right}: Use of the schema by the decoder.
    For clarity, the decoder outputs tokens rather than grammar rules.}
    \label{fig:schema_example}
\end{figure*}

where $\glegal$, $\vlegal$ are the number of legal rules (according to the grammar) that can be chosen at time step $j$ for schema-independent and schema-specific rules respectively. The score $s^\text{glob}_j$ is computed with a feed-forward network, and the score $s^\text{loc}_j$ is computed for all legal schema items by multiplying the matrix $S_\text{link} \in \mathbb{R}^{\vlegal \times |x|}$, which contains the relevant linking scores $s_\text{link}(v, x_i)$, with the attention vector $a_j$. Thus, decoding unseen schema items is done by first attending to the question words, which are linked to the schema items. 
\section{Modeling Schemas with GNNs}
\label{sec:model}

Schema structure is informative for predicting the SQL query. Consider a table with two columns, where each is a foreign key to two other tables (\texttt{student\_semester} table in Figure~\ref{fig:schema_example}).
Such a table is commonly used for describing a many-to-many relation between two other tables, which affects the output query. We now show how we represent this information in a neural parser and use it to improve predictions.

At a high-level our model has the following parts (Figure~\ref{fig:schema_example}). (a) The schema is converted to a graph. (b) The graph is softly pruned conditioned on the input question. (c) A Graph neural network generates a representation for nodes that is aware of the global schema structure. (d) The encoder and decoder use the schema representation. We will now elaborate on each part.

\paragraph{Schema-to-graph}
To convert the schema $S$ to a graph (Figure~\ref{fig:schema_example}, left), we define the graph nodes as the schema items $\sV$. We add three types of edges: for each column $c_t$ in a table $t$, we add edges $(c_t, t)$ and $(t, c_t)$ to the edge set $\sE_\leftrightarrow$ (green edges). For each foreign-primary key column pair $(c_{t_1}, c_{t_2}) \in \sF$, we add edges $(c_{t_1}, c_{t_2})$ and $(t_1, t_2)$ to the edge set $\sE_\rightarrow$ and edges $(c_{t_2}, c_{t_1})$ and $(t_2, t_1)$ to $\sE_\leftarrow$ (dashed edges). Edge types are used by the graph neural network to capture different ways in which columns and tables relate to one another.

\paragraph{Question-conditioned relevance}
Each question refers to different parts of the schema, and thus, our representation should change conditioned on the question. For example, in Figure \ref{fig:schema_example}, the relation between the tables \texttt{student\_semester} and \texttt{program} is irrelevant. To model that, we re-use the distribution $p_\text{link}(\cdot)$ from \S\ref{sec:base_model}, and define a relevance score for a schema item $v$: 
$\rho_v = \max_i{p_\text{link}(v \mid x_i)}$ --- 
the maximum probability of $v$ for any word $x_i$. We use this score next to create a question-conditioned graph representation. Figure~\ref{fig:schema_example} shows relevant schema items in dark orange, and irrelevant items in light orange.

\paragraph{Neural graph representation}
To learn a node representation that considers its relevance score and the global schema structure, we use gated GNNs \cite{li2016gated}. Each node $v$ is given
an initial embedding conditioned on the relevance score: $h^{(0)}_v = r_v \cdot \rho_v$. 
We then apply the GNN recurrence for $L$ steps. At each step, each node re-computes its representation based on the representation of its neighbors in the previous step:
$$
    a_v^{(l)} = \sum_{\text{type} \in \{\rightarrow, \leftrightarrow\}}
    \sum_{(u, v) \in \sE_{\text{type}}}  W_{\text{type}} h_{u}^{l-1} + b_{\text{type}} \label{line:gnn},
$$
and then $h_v^{(l)}$ is computed as following, using a standard GRU  \cite{cho-etal-2014-learning} update: $$h_v^{(l)} = \text{GRU}(h_v^{(l-1)}, a_v^{(l)})$$ (see \citet{li2016gated} for further details).

We denote the final representation of each schema item after $L$ steps by $\varphi_v=h_v^{(L)}$. 
We now show how this representation is used by the parser.

\paragraph{Encoder}
In \S\ref{sec:base_model}, 
a weighted average over schema items $l_i$ was concatenated to every word $x_i$. 
To enjoy the schema-aware representations, we compute $l^\varphi_i = \sum_\tau \sum_{v \in \sV_\tau} \varphi_v p_{\text{link}}(v \mid x_i)$, which is identical to $l_i$, except $\varphi_v$ is used instead of $r_v$. We concatenate $l^\varphi_i$ to the output of the encoder $h_i$, so that each word is augmented with the graph structure around the schema items it is linked to.

\paragraph{Decoder}
As mentioned (\S\ref{sec:base_model}), when a schema item $v$ is decoded, the input in the next time step is its type $\tau(v)$. A first change is to replace $\tau(v)$ by $\varphi_v$, which has knowledge of the structure around $v$.
A second change is a self-attention mechanism that links to the schema, which we describe next.

When scoring a schema item, its score should depend on its relation to previously decoded schema items. E.g., in Figure \ref{fig:schema_example}, once the table \texttt{semester} has been decoded, it is likely to be joined to a related table. We capture this intuition with a self-attention mechanism. 

For each decoding step $j$, we denote by $u_j$ the hidden state of the decoder, and by $\hat{J} = (i_1, \dots, i_{|\hat{J}|})$ the list of time steps before $j$ where a schema item has been decoded. We define the matrix $\hat{U} \in \mathbb{R}^{d \times |\hat{J}|} = [u_{i_1}, \dots, u_{i_{|\hat{J}|}}]$, which concatenates the hidden states from all these time steps. We now compute a self-attention distribution over these time steps, and score schema items based on this distribution (Figure~\ref{fig:schema_example}, right):
\begin{align*}
    \hat{a}_j &= \softmax(\hat{U}^T u_j) \quad \in \mathbb{R}^{|\hat{J}|}, \\
    s^{\text{att}}_j &= \hat{a}_j S^\text{att}, \\
    p_j &= \softmax([s^\text{glob}_j;s^\text{loc}_j + s^\text{att}_j]),
\end{align*}

where the matrix $S^\text{att} \in \mathbb{R}^{|\hat{J}| \times \vlegal}$ computes a similarity between
schema items that were previously decoded, and schema items that are legal according to the grammar: $S^\text{att}_{v_1, v_2} = F(\varphi_{v_1})^\top F(\varphi_{v_2})$, where $F(\cdot)$ is a feed-forward network. Thus, the score of a schema item increases, if substantial attention is placed on schema items to which it bears high similarity. 

\paragraph{Training} We maximize the log-likelihood of the gold sequence during training, and use beam-search (of size 10) at test time, similar to \citealt{krishnamurthy2017neural} and prior work. We run the GNN for $L=2$ steps.
\section{Experiments and Results}
\label{sec:experiments}

\paragraph{Experimental setup}
We evaluate on 
\spider{} \cite{yu2018spider}, which contains 7,000/1,034/2,147 train/development/test examples.

We pre-process examples to remove table aliases (\texttt{AS T1/T2/...}) from the queries and use the explicit table name instead (i.e. we replace \texttt{T1.col} with \texttt{table1\_name.col}), as in the majority of the cases ($>$ 99\% in \spider{}) these aliases are redundant. In addition, we add a table reference to all columns that do not have one (i.e. we replace \texttt{col} with \texttt{table\_name.col}).

We use the official evaluation script from \spider{} to ≈compute accuracy, i.e., whether the predicted query is equivalent to the gold query.

\paragraph{Results}
Our full model (\textsc{GNN}) obtains 39.4\% accuracy on the test set, substantially higher than prior state-of-the-art (\textsc{SyntaxSQLNet}), which is at 19.7\%. Removing the GNN from the parser (\textsc{No GNN}), which results in the parser of \newcite{krishnamurthy2017neural}, augmented with a grammar for SQL, obtains an accuracy of 33.8\%, showing the importance of encoding the schema structure.

\begin{table}[t]
\centering
{\small
\begin{tabular}{|l l l l|}
\hline
Model                     & Acc. & \textsc{Single} & \textsc{Multi} \\ \hline
\textsc{SQLNet}                    & 10.9\%   &  13.6\% &    3.3\%     \\ 
\textsc{SyntaxSQLNet}              & 18.9\% & 23.1\% & 7.0\%           \\ \hline
\textsc{No GNN}                    & 34.9\%  & 52.3\% & 14.6\% \\ 
\textsc{\textbf{GNN}}         & \textbf{40.7\%}  & 52.2\% & \textbf{26.8\%}    \\ 
- \textsc{No Self Attend}                       & 38.7\%  & \textbf{54.5\%} & 20.3\%     \\ 
- \textsc{Only self attend}                       & 35.9\%   & 47.1\%  & 23.0\%       \\ 
- \textsc{No Rel.} & 37.0\%  & 50.4\% & 21.5\%         \\ 
\hline
\textsc{GNN Oracle Rel.}  & 54.3\%  & 63.5\% & 43.7\%         \\ 
\hline
\end{tabular}
}
\caption{Development set accuracy for all models.}
\label{tab:results}
\end{table}

Table~\ref{tab:results} shows results on the development set for baselines and ablations. The first column describes accuracy on the entire dataset, and the next two columns show accuracy when partitioning examples to queries involving only one table (\textsc{Single}) vs. more than one table (\textsc{Multi}).

\text{GNN} dramatically outperforms previously published baselines \textsc{SQLNet} and \textsc{SyntaxSQLNet}, and improves the performance of \textsc{No GNN} from 34.9\% to 40.7\%. Importantly, using schema structure specifically improves performance on questions with multiple tables from 14.6\% to 26.8\%.

We ablate the major novel components of our model to assess their impact. First, we remove the self-attention component (\textsc{No Self Attend}). We observe that performance drops by 2 points, where \textsc{Single} slightly improves, and \textsc{Multi} drops by 6.5 points. Second, to verify that improvement is not only due to self-attention, we ablate all other uses of the GNN. Namely, We use a model identical to \textsc{No GNN}, except it can access the GNN representations through the self-attention (\textsc{Only Self Attend}). We observe a large drop in performance to 35.9\%, showing that all components are important. Last, we ablate the relevance score by setting $\rho_v = 1$ for all schema items (\textsc{No Rel.}). Indeed, accuracy drops to 37.0\%.

To assess the ceiling performance possible with a perfect relevance score, we run an oracle experiment, where we set $\rho_v = 1$ for all schema items that are in the gold query, and $\rho_v = 0$ for all other schema items (\textsc{GNN Oracle Rel.}). We see that a perfect relevance score substantially improves performance to 54.3\%, indicating substantial headroom for future research.

\noindent
\textbf{\texttt{join} analysis}
For any model, we can examine the proportion of predicted queries with a \texttt{join}, where the structure of the \texttt{join} is ``bad": (a) when the \texttt{join} condition clause uses the same table twice (\texttt{ON t1.column1 = t1.column2}), and (b) when the joined table are not connected through a primary-foreign key relation.

We find that \textsc{No GNN} predicts such \texttt{join}s in 83.4\% of the cases, while \textsc{GNN} does so in only 15.6\% of cases. 
When automatically omitting from the beam candidates where condition (a) occurs, \textsc{No GNN} predicts a ``bad" \texttt{join} in 14.2\% of the cases vs.  4.3\% for \textsc{GNN} (total accuracy increases by 0.3\% for both models). As an example, in Figure \ref{fig:schema_example}, 
$s^{\text{loc}}_j$ scores the table \texttt{student} the highest, although it is not related to the previously decoded table \texttt{semester}. Adding the self-attention score $s^{\text{att}}_j$ corrects this and leads to the correct \texttt{student\_semester}, probably because the model learns to prefer connected tables.

\section{Conclusion}

We present a semantic parser that encodes the structure of the DB schema with a graph neural network, and uses this representation to make schema-aware decisions both at encoding and decoding time. We demonstrate the effectivness of this method on \spider{}, a dataset that contains complex schemas which are not seen at training time, and show substantial improvement over current state-of-the-art.

\section*{Acknowledgments}
We thank Kevin Lin and Mark Neumann from Allen Institute for Artificial Intelligence for their help with the SQL grammar. This research was supported by Facebook. This work was completed in partial fulfillment for the Ph.D degree of the first author.

\bibliography{all}
\bibliographystyle{acl_natbib}

\appendix

\clearpage
\onecolumn

\begin{table}
\centering
\footnotesize
\begin{tabular}{|>{\hspace{0pt}}p{0.996\linewidth}|}
\hline
Question: List the names of poker players ordered by the final tables made in ascending order.                                                  \\
\hline
poker player: ['poker player id', 'people id', 'final table made', 'best finish', 'money rank', 'earnings'] \\people: ['people id', 'nationality', 'name', 'birth date', 'height']                                                                                                                                                      \\
\hline
\correct{\textsc{GNN}:~\texttt{SELECT people.name FROM poker\_player JOIN people ON poker\_player.people\_id = \newline people.people\_id ORDER BY poker\_player.final\_table\_made ASC}}              \\
\hline
\notcorrect{\textsc{NoGNN}:~\texttt{SELECT people.name FROM people ORDER BY people.name DESC}}                                                                                              \\

\hline \multicolumn{1}{>{\hspace{0pt}}p{0.998\linewidth}}{}\\\hline
Question: Which city has the most frequent destination airport?                                                  \\
\hline
airlines: ['airline id', 'airline name', 'abbreviation', 'country'] \\airports: ['city', 'airport code', 'airport name', 'country', 'country abbrev'] \\ flights: ['airline', 'flight number', 'source airport', 'destination airport']                                                                                                                                                       \\
\hline
\correct{\textsc{GNN}:~\texttt{SELECT airports.city FROM airports JOIN flights ON airports.airportcode = \newline flights.destairport GROUP BY airports.city ORDER BY count (*) DESC LIMIT 1}}              \\
\hline
\notcorrect{\textsc{NoGNN}:~\texttt{SELECT airports.city FROM airports GROUP BY airports.city ORDER BY count (*) \newline DESC LIMIT 1}}                                                                                              \\

\hline \multicolumn{1}{>{\hspace{0pt}}p{0.998\linewidth}}{}\\\hline
Question: What are the names of airports in Aberdeen?                                                  \\
\hline
airlines: ['airline id', 'airline name', 'abbreviation', 'country'] \\airports: ['city', 'airport code', 'airport name', 'country', 'country abbrev'] \\ flights: ['airline', 'flight number', 'source airport', 'destination airport']                                                                                                                                                       \\
\hline
\notcorrect{\textsc{GNN}:~\texttt{SELECT airports.airportname FROM airports WHERE airports.airportname = \newline "value"}}              \\
\hline
\correct{\textsc{NoGNN}:~\texttt{SELECT airports.airportname FROM airports WHERE airports.city = "value"}}                                                                                              \\

\hline \multicolumn{1}{>{\hspace{0pt}}p{0.998\linewidth}}{}\\\hline
Question: List the language used least number of TV Channel. List language and number of TV Channel.                                                  \\
\hline
tv channel: ['id', 'series name', 'country', 'language', 'content', 'pixel aspect ratio par', 'hight definition tv', 'pay per view ppv', 'package option'] \\ tv series: ['id', 'episode', 'air date', 'rating', 'share', '18 49 rating share', 'viewers m', 'weekly rank', 'channel'] \\ cartoon: ['id', 'title', 'directed by', 'written by', 'original air date', 'production code', 'channel']                                                                                                                                                       \\
\hline
\correct{\textsc{GNN}:~\texttt{SELECT tv\_channel.language, count (*) FROM tv\_channel GROUP BY \newline tv\_channel.language ORDER BY count (*) ASC LIMIT 1}}              \\
\hline
\notcorrect{\textsc{NoGNN}:~\texttt{SELECT tv\_channel.language, count (*) FROM tv\_channel GROUP BY \newline tv\_channel.language}}                                                                                              \\

\hline \multicolumn{1}{>{\hspace{0pt}}p{0.998\linewidth}}{}\\\hline
Question: Find the type and weight of the youngest pet.                                                  \\
\hline
student: ['student id', 'last name', 'first name', 'age', 'sex', 'major', 'advisor', 'city code'] \\has pet: ['student id', 'pet id']\\pets: ['pet id', 'pet type', 'pet age', 'weight']                                                                                                                                                       \\
\hline
\correct{\textsc{GNN}:~\texttt{SELECT pets.pettype, pets.weight FROM pets ORDER BY pets.pet\_age LIMIT 1}}              \\
\hline
\notcorrect{\textsc{NoGNN}:~\texttt{SELECT pets.pettype, pets.weight, pets.pet\_age FROM pets}}                                                                                              \\

\hline \multicolumn{1}{>{\hspace{0pt}}p{0.998\linewidth}}{}\\\hline
Question: List each owner's first name, last name, and the size of his for her dog.                                                  \\
\hline
breeds: ['breed code', 'breed name'] \\charges: ['charge id', 'charge type', 'charge amount']\\sizes: ['size code', 'size description']\\treatment types: ['treatment type code', 'treatment type description']\\owners: ['owner id', 'first name', 'last name', 'street', 'city', 'state', 'zip code', 'email address', 'home phone', 'cell number']\\dogs: ['dog id', 'owner id', 'abandoned yes or no', 'breed code', 'size code', 'name', 'age', 'date of birth', 'gender', 'weight', 'date arrived', 'date adopted', 'date departed']\\professionals: ['professional id', 'role code', 'first name', 'street', 'city', 'state', 'zip code', 'last name', 'email address', 'home phone', 'cell number']\\treatments: ['treatment id', 'dog id', 'professional id', 'treatment type code', 'date of treatment', 'cost of treatment']                                                                                                                                                       \\
\hline
\correct{\textsc{GNN}:~\texttt{SELECT owners.first\_name, owners.last\_name, dogs.size\_code FROM dogs JOIN \newline owners ON dogs.owner\_id = owners.owner\_id}}              \\
\hline
\notcorrect{\textsc{NoGNN}:~\texttt{SELECT owners.first\_name, owners.last\_name, count (*) FROM dogs JOIN owners \newline ON dogs.owner\_id = owners.owner\_id GROUP BY owners.owner\_id}}                                                                                              \\

\hline \multicolumn{1}{>{\hspace{0pt}}p{0.998\linewidth}}{}\\\hline
Question: What is the number of carsw ith over 6 cylinders?                                                  \\
\hline
continents: ['cont id', 'continent'] \\countries: ['country id', 'country name', 'continent']\\car makers: ['id', 'maker', 'full name', 'country']\\model list: ['model id', 'maker', 'model']\\car names: ['make id', 'model', 'make']\\cars data: ['id', 'mpg', 'cylinders', 'edispl', 'horsepower', 'weight', 'accelerate', 'year']                                                                                                                                                       \\
\hline
\correct{\textsc{GNN}:~\texttt{SELECT count (*) FROM cars\_data WHERE cars\_data.cylinders > "value"}}              \\
\hline
\notcorrect{\textsc{NoGNN}:~\texttt{SELECT count (*) FROM model\_list JOIN cars\_data ON cars\_data.id = \newline cars\_data.id WHERE cars\_data.cylinders > "value"}}                                                                                              \\

\hline
\end{tabular}
\caption{We randomly select query outputs from \textsc{No-GNN} and \textsc{GNN} in cases where one of the models was correct and the other made an error. We show questions, schemas and queries. Correct answers are marked in blue.}
\label{tab:samples}
\end{table}

\twocolumn

\clearpage

\label{app:outputs_examples}






\end{document}